\documentclass[letterpaper]{article}

\PassOptionsToPackage{numbers}{natbib}

\usepackage[final]{neurips_2021}

\usepackage[utf8]{inputenc} % allow utf-8 input
\usepackage[T1]{fontenc}    % use 8-bit T1 fonts
\usepackage{hyperref}       % hyperlinks
\usepackage{url}            % simple URL typesetting
\usepackage{booktabs}       % professional-quality tables
\usepackage{amsfonts}       % blackboard math symbols
\usepackage{nicefrac}       % compact symbols for 1/2, etc.
\usepackage{microtype}      % microtypography
\usepackage{xcolor}         % colors
\usepackage{graphicx}
\usepackage{stackrel}
\usepackage{mathtools}
\usepackage{amsthm}   % proof
\usepackage{bm}
\usepackage{subcaption}
\usepackage{adjustbox}
\usepackage{caption}       % for captions in minipage

\usepackage{algorithm}
\usepackage{algorithmicx}
\usepackage[noend]{algpseudocode}

\usepackage{amsmath}
\DeclareMathOperator*{\argmax}{argmax}
\DeclareMathOperator*{\argmin}{argmin}

\newcommand{\expectsub}[2]{{\mathcal E}_{#1}\left[ #2 \right]}

\newcommand{\prob}{\operatorname{Pr}\probarg}

\DeclarePairedDelimiterX{\expectarg}[1]{{[}}{{]}}{%
  \ifnum\currentgrouptype=16 \else\begingroup\fi
  \activatebar#1
  \ifnum\currentgrouptype=16 \else\endgroup\fi
}
\DeclarePairedDelimiterX{\probarg}[1]{(}{)}{%
  \ifnum\currentgrouptype=16 \else\begingroup\fi
  \activatebar#1
  \ifnum\currentgrouptype=16 \else\endgroup\fi
}
\newcommand{\innermid}{\nonscript\;\delimsize\vert\nonscript\;}
\newcommand{\activatebar}{%
  \begingroup\lccode`\~=`\|
  \lowercase{\endgroup\let~}\innermid
  \mathcode`|=\string"8000
}

\newtheorem{theorem}{Theorem}
\newtheorem{lemma}{Lemma}

\newcommand{\aside}[1]{}            %  use to comment out something

\title{Diversity Enhanced Active Learning with Strictly Proper Scoring Rules}

\author{%
  Wei Tan\\
  Monash University\\\texttt{wei.tan2@monash.edu} 
     \and
  Lan Du\thanks{Corresponding author}\\
  Monash University\\\texttt{lan.du@monash.edu} 
   \and
  Wray Buntine\\
  Monash University\\\texttt{wray.buntine@monash.edu} 
}

\begin{document}

\maketitle

%% I incorporated David's stuff, as well as some BEMPS 
\begin{abstract}
  We study acquisition functions for active learning (AL) for text classification. 
  The Expected Loss Reduction (ELR) method focuses on a Bayesian estimate of the reduction in classification error,
  recently updated with Mean Objective Cost of Uncertainty (MOCU).  
  We convert the ELR framework to estimate the increase in (strictly proper) scores like log probability or negative mean square error, 
  which we call Bayesian Estimate of Mean Proper Scores (BEMPS\footnote{Our implementation of BEMPS can be downloaded from \url{https://github.com/davidtw999/BEMPS}.}).
  We also prove convergence results borrowing techniques used with MOCU.
  In order to allow better experimentation with the new acquisition functions, 
  we develop a complementary batch AL algorithm, 
  which encourages diversity in the vector of expected changes in scores for unlabelled data.  
  % Further, we develop a pool filtering approach that provides initial screening of the unlabelled pool to further speed up AL with large pools
  To allow high performance text classifiers, 
  we combine ensembling and dynamic validation set construction on pretrained language models.  
  Extensive experimental evaluation then explores how these different acquisition functions 
%   and the implementation speedups 
  perform.  
  The results show that the use of mean square error and log probability with BEMPS yields robust acquisition functions, which consistently outperform the others tested.
\aside{Moreover, our implementation allows this class of estimation-based acquisition functions to be scaled to large pool sizes of 100,000.}
\end{abstract}

\section{Introduction}

Classification has extensive uses and deep learning has substantially
improved its performance,
but a major hurdle for its use is the paucity of labelled or annotated data.
The data labelling process performed by domain experts is expensive and tedious to produce, 
especially in the medical field, 
where due to lack of expertise and privacy issues, annotation costs are time-consuming and expensive \cite{jimenez2018capsule}.
Active Learning  (AL) is an approach to speeding up learning by 
judiciously selecting data to be annotated \cite{settles2009active}.
AL is perhaps the simplest of all human-in-the-loop
learning approaches, 
because the only interaction is the (expert) human providing a class label,
yet for even in the simplest of tasks,
classification, a general theory of AL is not agreed on.  

Moreover, in practical situations, one needs to consider many issues when designing an AL system with deep learning:
the expense to retrain deep neural networks and
the use of validation data for training \cite{siddhant-lipton-2018-deep}, 
transformer language models \cite{DBLP:journals/corr/abs-2004-13138}, %\cite{10.1145/3373017.3373028} 
batch mode AL with diversity \cite{ren2020survey},
and consideration of expert capabilities  and costs  \cite{gao2020cost, zhdanov2019diverse}.
Also important in experimental work is the need for 
realistic labelling sizes,
with practitioners we work with saying expert annotation
%, so without the use of crowd-sourcing, 
may allow a budget of up to 1000
data points, rarely more.
Using large batch sizes (e.g., 1000 in \cite{ash2019deep,NEURIPS2019_84c2d486}) can thus be impractical.
\aside{
Moreover, one should also consider employing state of the art
semi-supervised learning methods which are intended for a 
similar scenario, that is, smaller amounts of labelled data with a large pool of unlabelled data.
}

Our fundamental research contribution is to suggest
what makes a good acquisition function for the uncertainty component, without any batching.
While there are many recent methods looking at the uncertainty diversity trade-off for
batch AL \cite{ren2020survey}, few recently have focused on understanding the uncertainty side alone.
A substantial advance is a recent theoretical framework, Mean Objective Cost of Uncertainty (MOCU) \cite{ZhaoICLR21} that provides a convergence proof.
% based around the Expected Loss Reduction (ELR) framework of Bayesian regret computed on classification errors \cite{RoyMcC2001}, that also provides a convergence proof.  
As our first contribution, we convert their expected loss reduction framework
to an AL model using strictly proper scoring rules or Bregman divergences \cite{doi:10.1198/016214506000001437} 
instead of classification errors.  MOCU required manipulations of the expected error,
resulting in weighted-MOCU (WMOCU), in order to achieve convergence and avoid 
getting stuck in error bands.  Strictly proper scoring rules naturally
avoid this problem by generalising expected errors to expected scores.
Using strictly proper scoring rules means better calibrated classifiers are rewarded.
The scoring rules go beyond simple minimum errors of WMOCU and can be adapted to different kinds of inference tasks (e.g., different utilities, precision-recall trade-offs, etc.).
This property is preferable and beneficial for applications such as medical domains where actual errors become less relevant for an inference task.

In order to evaluate the new acquisition functions we use text classification,
which is our target application domain.
For realistic evaluation, we want to use near state of the art systems,
which means using pretrained language models with validation sets \cite{sanh2019distilbert}, 
and neural network ensembles \cite{Lakshminarayanan2017}.  
Ensembling also doubles as a heuristic
technique to yield estimates of model uncertainty and posterior probabilities.
Coming up with a simple approach to combine ensembling and validations sets is our second research contribution.
The importance of these combinations for AL has been noted \cite{yuan-etal-2020-cold,schroder2020survey}.
For further batch comparisons,
% at larger scale, 
we then bring back diversity into the research,
suggesting a way to naturally complement our new family of acquisition functions with a method to achieve diversity, our third research contribution.
Extensive experiments with a comprehensive set of ablation studies
on four text classification datasets show that
our BEMPS-based AL model consistently outperforms recent techniques like WMOCU and BADGE,
although we explicitly exclude recent semi-supervised AL methods because they represent an unfair comparison against strictly supervised learning.
% We also implement our BEMPS on the open-source platform \footnote{\url{https://github.com/davidtw999/BEMPS}}.
% Our implementation of BEMPS can be downloaded from \url{https://github.com/davidtw999/BEMPS}.

\section{Related work}
The proposed BEMPS, as a general Bayesian model for acquisition functions, 
quantifies the model uncertainty using the theory of (strictly) proper scoring rules for categorical
variables \cite{doi:10.1198/016214506000001437}.
Thus, we first review existing acquisition functions (aka query strategies) 
proposed in some recent AL models that are most related to ours, and
refer interested readers to \cite{ren2020survey} for
a comprehensive discussion of recent strategies.
% \lan{
One common and simple heuristic often used is maximum entropy \cite{zhu2009active,zhu2012uncertainty,wang2014new,gal2017deep},
where one chooses samples that maximise the predictive entropy.
% }
Bayesian AL by disagreement (BALD) \cite{Houlsby2011} and its batch version (i.e.,
BatchBALD) \cite{KAJvAYG2019} instead compute the mutual information between the model predictions
and the model parameters, which indeed chooses samples that maximise the decrease in
expected entropy \cite{NEURIPS2019_84c2d486}.
% and since then, have been widely used in \cite{}.
Recently, inspired by the one-step-look-ahead strategy of ELR \cite{RoyMcC2001}, 
\citet{ZhaoICLR21} extended MOCU \cite{yoon2013quantifying} to WMOCU with a theoretical guarantee of convergence.
But WMOCU only applies to errors which are less relevant for some inference tasks.
BEMPS naturally extends those methods, focusing on Bayesian estimation
of uncertainty related to the model's expected proper score.

Ensemble methods are used to obtain better uncertainty estimates with deep learning, including Monte-Carlo dropout (MC-dropout) \cite{gal2016dropout,KAJvAYG2019,pop2020deep} and
deep ensembles \cite{Lakshminarayanan2017}.
More sophisticated techniques are being developed, for instance, MCMC, hybrid and deterministic approaches \cite{WilsonNIPS202,AshukhaLMV20}.
However, plain ensembling remains a  competitive and simple method for deep learning \cite{AshukhaLMV20}.

Query strategies considering just uncertainty do not always work well in a batch setting, due to the highly similar samples acquired in a batch \cite{WangZengmao2016Abal, ren2020survey,ma2021active}.
To overcome this problem, there have been many AL methods that
achieve batch diversity by acquiring samples that are both informative and diverse, such as \cite{Nguyen2004,yin2017deep, PengLiu2017ADLf,ash2019deep, yuan-etal-2020-cold,zhdanov2019diverse, shi-etal-2021-diversity}.
BADGE \cite{ash2019deep} and ALPS \cite{yuan-etal-2020-cold} 
are the two recent AL methods focusing on batch diversity. 
BADGE uses gradient embeddings of unlabelled samples as inputs of $k$-MEANS++ to select a set of diverse samples, which relies on fine-tuning pretrained language models.
Instead of using gradient embeddings,
ALPS uses surprisal embeddings computed from
the predictive word probabilities generated by a masked language model for cold-start AL.
Whereas our BEMPS computes for each unlabelled sample
a vector of expected change in the proper scores, relating directly to the performance.

Other recent batch AL approaches, WAAL \cite{pmlr-v108-shui20a} and VAAL \cite{Sinha_2019_ICCV}
% and MAL \cite{ebrahimi2020minimax} 
use semi-supervised learning, for instance WAAL's models are trained on the feature space obtained with the help of the unlabelled data.
% , and these models are reporting good performance.  
%%%%%%%%%%%%%%%%%%%%%%%%%%
% We exclude semi-supervised learning algorithms from our comparison and thus need to exclude these methods.
%%%%%%%%%%%%%%%%%%%%%
We exclude these methods from our comparison since they are semi-supervised learning algorithms.
Both Core-set \cite{sener2018active} and WAAL \cite{pmlr-v108-shui20a} also set up sophisticated
non-Bayesian cost functions using bounds to deal with the many unknown probabilities when trying
to optimise for a batch, whereas the Bayesian formula of BEMPS can be directly
estimated.  
% In reported experiments \cite{ebrahimi2020minimax}, VAAL seems 0.5-1.5\% better in accuracy than 
% Core-set which appears to be similar to Random.

In experimental work, 
the use of a validation set to train deep learning models in active learning is not uncommon, the goal of which is to avoid over-fitting and achieve early-stopping.
Some existing methods assume that there
is a large validation set available a prior 
\cite{ash2019deep, KAJvAYG2019, gal2017deep},
but means the cost of labelling the set is not factored into the  labelling budget.
We argue that the availability of a separate validation set is impractical
in real world AL scenarios.
% the labelling budget for this is not factored into the AL experiments. 
Although
\citet{yuan-etal-2020-cold} use fixed epochs to train the classifier without a validation set to save the cost, 
 the classifier could either be under-fit or over-fit.
 We instead use a dynamic approach to generate alternative validation sets
 from the ever increasing labelled pool after each iteration.

\section{Bayesian estimate of mean proper scores}
We first review the general Bayesian model for acquisition functions that includes ELR, MOCU and BALD. 
%as a special case. 
We then develop Bayesian Estimate of Mean Proper Scores (BEMPS), 
a new uncertainty quantification framework with a theoretical foundation  based on 
strictly proper scoring rules \cite{doi:10.1198/016214506000001437}.
It naturally extends ELR and BALD, focusing on the Bayesian estimation
of uncertainty related to the models expected performance.
% We first review the general Bayesian model for acquisition functions that includes MOCU and BALD as a special case.  We then develop a new variant based on 
% strictly proper scoring rules.

Suppose models of our interest are parameterised by parameters $\theta \in \Theta$,
$L$ indicates labelled data,
% Suppose we have labelled data $L$, and models we consider are parameterised by
% parameters $\theta \in \Theta$, 
probability of label $y$ for data $x$ is given by $\prob{y|\theta,x}$,
and $\prob{\cdot|\theta,x}$ presents a vector of label probabilities.
% Moreover, it is 
With a \emph{fully conditional model}, %so that 
the posterior of $\theta$
is unaffected by unlabelled data, which means $\prob{\theta|L,U}=\prob{\theta|L}$ 
for any unlabelled data $U$.
Moreover, 
we assume without loss of generality that this model family is well behaved in a statistical sense, 
so the model is identifiable.
The ``cost" of the posterior $\prob{\theta|L}$ can be measured
by some functional $Q(\prob{\theta|L})$, denoted $Q(L)$ for short, where $Q(L)\ge 0$ and $Q(L)=0$
when some convergence objective has been achieved.  For our model this is
when $\prob{\theta|L}$ has converged to a point mass at a single model.  
A suitable objective function is to measure the expected decrease in $Q(\cdot)$ 
due to acquiring the label for a data point $x$.
The corresponding acquisition function for AL is formulated 
\cite[Equation~(1)]{Houlsby2011} as
\begin{equation}
    \label{eq-dQ}
\Delta Q(x|L) = Q(L) - \expectsub{\prob{y|L,x}}{Q(L\cup \{(x,y)\})}\,,
\end{equation}
whereas for ELR the expression is split over an inequality sign \cite[Equation~(2)]{RoyMcC2001}.
It estimates how much the cost is expected to reduce when a new data point $x$ is acquired.  
Since the true label for the new data $x$ is unknown a prior, 
we have to use expected posterior proportions
from our model, $\prob{y|L,x}$ to estimate the likely label.
For BALD \cite{Houlsby2011} using Shannon's entropy, $Q_I(L) = I(\prob{\theta|L})$, which measures uncertainty in the parameter space and thus has no strong relationship to actual errors \cite{ZhaoICLR21}.
MOCU and ELR use a Bayesian regret given by the expected loss difference between the
optimal Bayesian classifier and the optimal classifier:
\begin{eqnarray}
\label{eq-q-mocu}
Q_{MOCU}(L) = \expectsub{\prob{x'}}{ 
          \min_{y'} (1-\prob{y'|L,x'}) -  \expectsub{\prob{\theta|L}}{\min_{y'} (1-\prob{y'|\theta,x'}) } }\,.
\end{eqnarray}

WMOCU uses a weighting function defined by Eq (11) in \cite{ZhaoICLR21} to have a more amenable definition of $\Delta Q(x|L)$ than the MOCU method. Although WMOCU guarantees $\Delta Q(x|L)$ converging to the optimal classifier (under minimum errors) according to the $Q(L)$ with the strictly concave function by Eq (15) in \cite{ZhaoICLR21}, the optimal approximation of the convergences can only be solved by controlling a hyperparameter of the weighting function manually. To allow theoretical guarantees of convergence under more general loss functions, we propose a different definition for $Q(L)$ based on strictly proper scoring rules.

\subsection{Strictly Proper Scoring Rules for Active Learning}
\label{sct:theory}

% Our acquisition model uses strictly proper scoring rules so first we introduce basic details about them.
A scoring rule assesses the quality of probabilistic prediction of categorical variables,
and is often used in training a classification algorithm.
For a model $\prob{y|\theta,x}$ with input data $x$ and if one observes label $y$,
the score is given by a function
$S(\prob{\cdot|\theta,x},y)$.  A strictly proper scoring rule has the behaviour that in the limit of
infinite labelled data $L_{n}$, as $n\rightarrow \infty$, the average score
$\frac{1}{n} \sum_{(x,y)\in L_n} S(\prob{\cdot|\theta,x},y)$ has a unique maximum for $\theta$ at the
``true" model (for our identifiable model family).  
The Savage representation \cite{doi:10.1198/016214506000001437} states that a strictly proper scoring rule for categorical variables 
takes the  form
%%  WRAY: the representation q(\cdot) is my bad shorthand for a vector
%%    so the second term is in fact a vector dot product, hence the dagger
 $S(q(\cdot),y) = G(q(\cdot)) + dG(q(\cdot))(\delta_y-q(\cdot))$ %^{\dagger}
 for a strictly convex function $G(\cdot)$ with subgradient $dG(\cdot)$.
 Note that the expectation of a scoring rule according to the supplied probability takes a simple form $\expectsub{q(y)}{ S(q(\cdot),y)}=\sum_y q(y) S(q(\cdot),y)= G(q(\cdot))$.
 
With strictly proper scoring rules, 
we develop a generalised class of acquisition functions built using the posterior (i.e., w.r.t. $\prob{\theta|L}$) expected difference between the score for the Bayes optimal classifier and the score for the ``true" model.
This is inherently Bayesian due to the use of $\prob{\theta|L}$.
 \begin{eqnarray} 
 Q_S(L) &=&  \expectsub{ \prob{x} \prob{\theta|L} }{
    \expectsub{ \prob{y|\theta,x} }{
      S(\prob{\cdot|\theta,x},y) - S(\prob{\cdot|L,x},y)  } }\label{eq-QSS}\\
      &=&  \expectsub{ \prob{x} \prob{\theta|L} }{
    B(\prob{\cdot|L,x},\prob{\cdot|\theta,x}) }\label{eq-QB}\\\label{eq-QS}
      &=&  
   \expectsub{ \prob{x}}{ \expectsub{ \prob{\theta|L} }{G(\prob{\cdot|\theta,x})}
    - G( \prob{\cdot|L,x})}\\
 \Delta Q_S(x|L) &=&\expectsub{ \prob{x'}}{
  \expectsub{ \prob{y|L,x} }{G( \prob{\cdot|L,(x,y),x'})}
  - G( \prob{\cdot|L,x'}) }
  \label{eq-DQS}
 \end{eqnarray}

The $Q_S(L)$ has three equivalent variations, one for an arbitrary strictly proper scoring rule $ S(q(\cdot),y)$ (Eq~\eqref{eq-QSS}),
one for a corresponding Bregman divergence $B(\cdot,\cdot)$ (Eq~\eqref{eq-QB})
and the
third for an arbitrary strictly convex function $G(\cdot)$
(Eq~\eqref{eq-QS}).
Their connections are given in \cite{doi:10.1198/016214506000001437}.

It is noteworthy that the acquisition function $\Delta Q_S(x|L)$ defined in Eq~\eqref{eq-DQS} 
is in a general form,
applicable to any strictly proper scoring function for categorical variables.
For instance, using a logarithmic scoring rule, popular for deep neural networks,
we have $S_{log}(q(\cdot),y) = \log q(y)$  and $G_{log}(q(\cdot)) = -I(q(\cdot))$.
Using the squared error scoring rule, known as a Brier score,
we have $S_{MSE}(q(\cdot),y)=-\sum_{\hat{y}} \left(q(\hat{y})-1_{y=\hat{y}}\right)^2 $  and $G_{MSE}(q(\cdot)) = \sum_y q(y)^2 -1$.  Combining these $G(\cdot)$ functions with Equation~\eqref{eq-DQS} yields two acquisition functions for the different scoring rules.
These, as well as the corresponding for BALD have some elegant properties, with proofs given in Appendix~A.
\begin{lemma}[Properties of scoring]\label{thm-nn}
In the context of a fully conditional classification model $\prob{y|\theta,x}$,
the $Q_I(L)$, $Q_{S}(L)$, $\Delta Q_I(x|L)$, $\Delta Q_{S}(x|L)$ as defined above are all non-negative.
\end{lemma}
Moreover, they guarantee learning will converge to the ``truth".
\begin{theorem}[Convergence of active learning]\label{thm-cnv}
We have a fully conditional classification model $\prob{y|\theta,x}$, for $\theta\in \Theta$ 
with finite discrete classes $y$ and input features $x$.
Moreover, there is a unique ``true" model parameter $\theta_r$ with which the data is generated, the prior distribution $p(\theta)$ satisfies $p(\theta_r)>0$,
and the model is identifiable.
Then for the AL
algorithm defined by the acquisition functions defined above of
$\Delta Q_{I}(x|L)$ or $\Delta Q_{S}(x|L)$ after being applied for $n$ steps gives labelled data $L_n$,
then $\lim_{n \rightarrow \infty} \Delta  Q_I(x|L_n) = 0$
and likewise for $Q_{S}(\cdot)$.
Moreover, $\lim_{n \rightarrow \infty}\prob{\theta|L_n}$
is a delta function at $\theta=\theta_r$ for data acquired by
both $\Delta Q_{I}(x|L)$ or $\Delta Q_{S}(x|L)$.
\end{theorem}
Finiteness and discreteness of $x$ is used to adapt results from \cite{ZhaoICLR21}  to show for all $x$ that $\Delta Q(x|L_n)\rightarrow 0$ as $n \rightarrow \infty$,
not an issue since real data is always finite.
Interestingly $\Delta Q_{I}(x|L)$, i.e., BALD,  achieves convergence too,
which occurs
because the model is identifiable and fully conditional, during AL we are free to choose $x$ values that would distinguish different parameter values $\theta$.
Full conditionality also supports BEMPS
because it means any inherent bias in the AL
selection is nullified with the use of the data distribution $\prob{x}$.
But it also means that the theory has not been shown to hold for
semi-supervised learning algorithms, where full conditionality does not apply.

%
% add some rebuttal here based R1.1, R2.2 and R4.2
%

Compared with BALD, MOCU and WMOCU, the advantage of using strictly proper scoring rules in BEMPS is that they generalise expected errors to expected scores,
which can be tailored for different inference tasks.
BALD uses mutual information to score samples based on how their labels could inform the true model parameter distribution,
which will be problematic if the uncertainty of model parameters has a reduced relationship to the classification performance. 
This is reflected by its poor performance in our experiments.
MOCU however has convergence issues as ELR, as pointed out by \citet{ZhaoICLR21}. 
Even though WMOCU can overcome the convergence issues, it is limited to the minimisation of expected errors.

In contrast, measuring the quality of predictive distributions via rewarding calibrated predictive distributions, scoring rules are in favour of adaptability.
For instance,  scoring rules can be developed 
\cite{doi:10.1198/016214506000001437} for some  different inference tasks, including Brier score,
%spherical score, 
logarithmic score, etc. 
Many loss functions used by neural networks, like cross-entropy loss, are indeed proper scoring rules \cite{Lakshminarayanan2017}.
% one can adapt a scoring rule depending on whether quantiles are to be estimated, or a particular set of utilities are to be used. One could also be Bayesian about it and stick with a log probability but this presupposes a particular model family. 
In most cases we can create a particular model to match just about any Bregman divergence (e.g., minimum squared errors is a Gaussian). In practice we can use robust models (e.g., a Dirichlet-multinomial rather than a multinomial, a negative binomial rather than a Poisson, Cauchy rather than Gaussian) in our log probability. 
% So being Bayesian suggests using a robust model with a log probability score.
For the particular cases we used, minimising Brier score gets the probability right in a least squares sense, i.e., minimising
the squared error between the predictive probability and the one-shot label representation,
which pays less attention to very low probability events.
Meanwhile, log probability gets the probability scales right, paying attention to all events. 
% Brier score is known to work well with minimum errors tasks and non-extreme utility tasks.

\subsection{Acquisition Algorithms with Enhanced Batch Diversity}
\label{ssct:algos}

% \begin{algorithm}[!t]
% \small
%   \caption{Estimating point-wise $\Delta Q(x|L,x')$ with Equation~\eqref{eq-DQS} }\label{alg-qrx}
%   \begin{algorithmic}[1]\small
%     \Require unlabelled data point $x$, existing labelled data $L$, estimation point $x'$ 
%         \Require model/network ensemble $\Theta^E=\{\theta_1,...,\theta_E\}$ built from labelled data $L$,
%     \Require strictly convex function $G(\cdot)$ taking as input a probability density over $y$ 
    
%     \State $Q=0$
%     \State $qx(\cdot) = \sum_{\theta\in \Theta^E} \prob{\theta|L)}
%               \prob{ \cdot|\theta,x} $
%     \For{$y$}   
%         \State $q(\cdot) = \sum_{\theta\in \Theta^E} \prob{\theta|L,(x,y)}
%               \prob{ \cdot|\theta,x'} $
%         \State $Q ~+\!\!= qx(y)G(q(\cdot))$
%     \EndFor
%     \State $q(\cdot) = \sum_{\theta\in \Theta^E} \prob{\theta|L}
%               \prob{ \cdot|\theta,x'} $    
%   \State $Q~ -\!\!= G(q(\cdot))$
%     \State \Return $Q$
%   \end{algorithmic}
% \end{algorithm}

\begin{figure}[!t]\vspace*{-18pt}
\begin{minipage}{0.5\textwidth}
\vspace*{0pt}
\begin{algorithm}[H]\small
\small
  \caption{Estimating point-wise $\Delta Q(x|L,x')$ with Equation~\eqref{eq-DQS} }\label{alg-qrx}
  \begin{algorithmic}[1]\small
      \vspace{2.8mm}
    \Require unlabelled data point $x$, existing labelled data $L$, estimation point $x'$ 
          \vspace{1mm}
        \Require model/network ensemble $\Theta^E=\{\theta_1,...,\theta_E\}$ built from labelled data $L$,
              \vspace{1mm}
    \Require strictly convex function $G(\cdot)$ taking as input a probability density over $y$ 
          \vspace{1mm}
    \State $Q=0$
            \vspace{1mm}
    \State $qx(\cdot) = \sum_{\theta\in \Theta^E} \prob{\theta|L)}
              \prob{ \cdot|\theta,x} $
                      \vspace{1mm}
    \For{$y$}   
            \vspace{1mm}
        \State $q(\cdot) = \sum_{\theta\in \Theta^E} \prob{\theta|L,(x,y)}
              \prob{ \cdot|\theta,x'} $
                      \vspace{1mm}
        \State $Q ~+\!\!= qx(y)G(q(\cdot))$
                \vspace{1mm}
    \EndFor
            \vspace{1mm}
    \State $q(\cdot) = \sum_{\theta\in \Theta^E} \prob{\theta|L}
              \prob{ \cdot|\theta,x'} $   
                      \vspace{1mm}
   \State $Q~ -\!\!= G(q(\cdot))$
           \vspace{1mm}
    \State \Return $Q$
    \vspace{2.8mm}
  \end{algorithmic}
\end{algorithm}
% \begin{algorithm}[H]\small
%   \captionof{algorithm}{Estimate of $\argmax_{x\in U} \Delta Q(x|L)$ }\label{alg-qr}
%   \begin{algorithmic}[1]
%     \Require unlabelled pool $U$, estimation pool $X$
%   % \State pre-compute $\prob{\theta|L}$ and $\prob{\theta|L,(x,y)}$ for each $y$ 
%   % and $\theta\in \Theta_E$
%   \For{$x \in U$}
%     \State $Q_x=0$
%     \For{$x' \in X$}
%          \State $Q_x ~ +\!\!= \Delta Q(x|L,x')$
%     \EndFor \EndFor
%     \State \Return $\argmax_{x \in U} Q_x$ 
%     % \david{Is here argmax($Q/|x'|$)}  -- originally was just giving for x, changed to getting maximum
%   \end{algorithmic}
% \end{algorithm}
\end{minipage}\hspace*{0.02\textwidth}
\begin{minipage}{0.48\textwidth}\vspace*{0pt}
\begin{algorithm}[H]\small
  \captionof{algorithm}{Estimate of $\argmax_{x\in U} \Delta Q(x|L)$ }\label{alg-qr}
  \begin{algorithmic}[1]
    \Require unlabelled pool $U$, estimation pool $X$
   % \State pre-compute $\prob{\theta|L}$ and $\prob{\theta|L,(x,y)}$ for each $y$ 
   % and $\theta\in \Theta_E$
   \For{$x \in U$}
    \State $Q_x=0$
    \For{$x' \in X$}
         \State $Q_x ~ +\!\!= \Delta Q(x|L,x')$
    \EndFor \EndFor
    \State \Return $\argmax_{x \in U} Q_x$ 
    % \david{Is here argmax($Q/|x'|$)}  -- originally was just giving for x, changed to getting maximum
  \end{algorithmic}
\end{algorithm}
\vspace{-7mm}
\begin{algorithm}[H]\small
  \captionof{algorithm}{Finding a diverse batch}\label{alg-qrd}
  \begin{algorithmic}[1]
    \Require unlabelled pool $U$,  batch size $B$
        \Require estimation pool $X$,  top fraction $T$
    \State  $\forall_{x \in U} Q_{x} = 0$
    \For{$x \in U$, $x' \in X$}
         \State $Q_{x}~ +\!\!= vec_{x,x'} = \Delta Q(x|L,x')$
    \EndFor
   % \algorithmiccomment \textcolor{red}{what is $k$ for the topk operation?}\david{topk means top k value (20\%) which is equivalent to 0.2 in this case}
    \State $V \leftarrow topk(Q,T*|U|)$
    % Select top $k$\% vec based on Q value    %\algorithmiccomment returns clustering of points in $U$
    \State $batch = \emptyset$
    \State $centroids$ = $k$-Means centers $(vec_{x\in V}, B)$
    \For{$c \in centroids$}
        % \State $b = arg min_{x \in V}||c - vec_{x}||$
         \State $batch ~\cup\!\!= \{ \argmin_{x \in V}||c - vec_{x}|| \}$
    \EndFor
     \State \Return $batch$
  \end{algorithmic}
\end{algorithm}
\end{minipage}
\vspace{-5mm}
\end{figure}

Algorithm~\ref{alg-qr} gives an implementation of BEMPS for 
an arbitrary strictly convex function $G(\cdot)$, returning the data point with the best estimated measure.
To work with a Bregman divergence or proper score, the corresponding strictly convex function $G(\cdot)$ should first be derived.
When $G(\cdot)$ is negative entropy, we call this CoreLog and
when $G(\cdot)$ is the  sum of squares we call this CoreMSE,
corresponding to the log probability or Brier scoring rules respectively.
Algorithm~\ref{alg-qr} calls Algorithm~\ref{alg-qrx} to get the estimation 
at test point $x'$, which implements the function inside $\expectsub{ \prob{x'}}{\cdot}$ in Equation~\eqref{eq-DQS}.
Note $\prob{\theta|L,(x,y)}$ is computed from $\prob{\theta|L}$ using
Bayes theorem.
Both Algorithms~\ref{alg-qr} and~\ref{alg-qrd} use a fixed {\it estimation pool}, $X$, a fixed random subset of the initial unlabelled data
used to estimate expected values $\expectsub{ \prob{x'}}{\cdot}$.
Algorithm~\ref{alg-qrd} returns $B$ data points representing a batch with enhanced diversity:
it first calls Algorithm~\ref{alg-qrx} to get, 
for each data point $x$ in the unlabelled pool, 
the vector of expected changes in score values over the estimation pool.
Thus, this vector conveys information about  uncertainty directly
related to the change in score due to the addition of $x$.
While the gradient embedding used in \cite{ash2019deep}
represents a sample's impact on the model, our vector represents a sample's impact on the
mean proper score.
Concurrently Algorithm~\ref{alg-qrd} computes the estimate of $\Delta Q(x|L)$ for these same $x$s.
The top $T$\% of scoring data $x$ are then clustered with $k$-Means and a representative
of each cluster closest to the cluster mean is returned.
This $k$-Means selection process tends to generate a diverse batch
of high-scoring samples.
The intuition is that 1) only higher scoring data $x$ should appear in a batch;
2) those clusters capture the pattern of expected changes in score values
deduced by samples in the unlabelled pool, where
the samples with a similar mean change in score values are grouped together;
% in the same cluster;
3) samples in the same cluster can affect the 
learning similarly, so should not co-occur in a batch.

\begin{table}[!t]
  \caption{Datasets and the used language model}
  \label{tab:table1}
  \centering\small
  \begin{tabular}{llllll}
    \toprule
    Dataset & Unlabelled/Test sizes &  \# Classes & Lang.~Model & Initial labelled size  \\
    \midrule
    AG NEWS  & 120,000 / 7,600 & 4 & DistilBERT  &   26  \\
    PUBMED 20K RCT & 15,000 / 2,500 & 5 & DistilBERT & 26    \\
    IMDB     &  25,000 / 25,000 & 2 & DistilBERT       & 26 \\
    SST-5     & 8544 / 2210 & 5 & DistilBERT       & 26   \\
    \bottomrule
  \end{tabular}
  \vspace{-5mm}
\end{table}

\section{Experiments}

We demonstrate the efficacy of BEMPS via evaluating the performance of  CoreMSE and CoreLog
as its two examples on various benchmark datasets for text classification.
These two acquisition functions were compared with recent techniques for AL. 
% select the small number of samples from the acquired sample pool as the validation set on each AL acquisition iteration. We report both F1 and accuracy scores for the performance. 
% Meanwhile, a pretrained DistilBERT \cite{sanh2019distilbert} was used across all our experiments as the backbone text classifier.
% even though BEMPS can also be applied to other classification tasks, like image classification.

\textbf{Datasets}
% \autoref{tab:table1} contains the summary of the datasets and the model used in the experiment. 
We used {\it four benchmark text datasets} for three different classification tasks: topic classification, sentence classification, and sentiment analysis, as shown in \autoref{tab:table1}.
The AG NEWS for topic classification contains 120K texts of four balanced classes \cite{zhang2015character}. 
% To study the advantages of evaluating active learning algorithms in more practical scenarios, 
The PUBMED 20k was used for sentence classification \cite{dernoncourt2017pubmed},
which contains about 20K medical abstracts with five categories. 
This dataset is imbalanced. 
For sentiment analysis, we used both the SST-5 and the IMDB datasets. 
SST-5 contains 11K sentences extracted from movie reviews with five imbalanced sentiment labels \cite{socher2013recursive}, 
and IMDB contains 50K movie reviews with two balanced classes \cite{maas2011learning}.

\textbf{Baselines}
% We evaluate the performance of CoreMSE and CoreLog against various uncertainty and diversity active learning methods, including
% We considered Max-Entropy \cite{yang2016active}, BALD \cite{Houlsby2011}, MOCU \cite{ZhaoICLR21}, WMOCU \cite{ZhaoICLR21} and BADGE \cite{ash2019deep} together with a random baseline. 
%%%%%%%%%%%%%%%%%%%%%%%%%%%%%%%%%%%%%%%%%%%%%%%%%%%%%%%%%%%%%%%%%%%%%% comment out 793, and uncomment 795
We considered Max-Entropy \cite{yang2016active}, BALD \cite{Houlsby2011}, MOCU \cite{ZhaoICLR21}, WMOCU \cite{ZhaoICLR21}, BADGE \cite{ash2019deep} and ALPS \cite{yuan-etal-2020-cold} together with a random baseline. 
%%%%%%%%%%%%%%%%%%%%%%%%%%%%%%%%%%%%%%%%%%%%%%%%%%%%%%%%%%%%%%%%%%%%%%
% We exclude semi-supervised learning approaches such as VAAL and MAL, which mix in fundamentally superior text classification indirectly via semi-supervision, and thus corrupt the experiments.
Max-Entropy and BALD are earlier uncertainty-based methods. 
% Max-Entropy considers the relative uncertainty between different classes and measures the highest entropy of the predictive distribution \cite{lewis1994sequential, wang2014new}. 
% BALD 
% % focus on both relative uncertainty and model uncertainty. It 
% selects the data point that maximises the mutual information based on how well the label would 
% improve the model parameter distribution. % \cite{Houlsby2011}. 
% To extend the knowledge of the uncertainty on classification tasks, 
%MOCU, WMOCU and BADGE are the three recent AL methods.
%%%%%%%%%%%%%%%%%%%%%%%%%%%%%%%%%%%%%%%%%%%%%%%%%%%%%%%%%%%%%%%%%%%%%%
 MOCU, WMOCU, BADGE and ALPS are the four recent AL methods.
%%%%%%%%%%%%%%%%%%%%%%%%%%%%%%%%%%%%%%%%%%%%%%%%%%%%%%%%%%%%%%%%%%%%%%
% MOCU and WMOCU focus on the Bayesian estimation of reduction in the uncertainty, similar to our BEMPS. 
% that affects the classification error based
% on a one-step-look-ahead strategy, the idea of which is in the same spirit of our BEMPS.
% Based on a one-step-look-ahead strategy, the acquisition method maximises the expected reduction given a new data point. 
% In a naive batch implementation of these uncertainty methods, the selected data points are sometimes similar, with little new information about the model. 
% In order to maximise batch diversity, 
% BADGE 
% % uses gradient embeddings of unlabelled samples as inputs of $k$-MEANS++ to select a batch of samples, 
% maximises batch diversity with a $k$-MEANS++ algorithm.
% Those are the AL methods that are most related to ours, but almost all of them are implemented for image classification.
Following their originally published algorithms,
we re-implemented them for text classification tasks, with the same backbone classifier for fair comparisons.
Note that the hypothetical labels used in BADGE were computed with ensembles.  
% provides the data representation that contains model confidence to improve over uncertainty based methods. 
% \wray{have to brielfy describe *your* implementation of BADGE.}
%%  next sentence is results, Wray
% Our experiments demonstrate how CoreMSE and CoreLog perform on the text datasets against the uncertainty base methods on a single query data selection. 
 
% To ensure a comprehensive comparison between all methods, we evaluate them in both non-batch and batch versions of  active learning,
% however, for MOCU and WMOCU we adapt our batch Algorithm~\ref{alg-qrd} using their own pointwise error estimates.
% We also show results using the Random query strategy, a naive baseline for randomly selecting unlabelled pool data.

% \subsection{Text classification, hyper-parameter settings and implementation}
\textbf{Experiment setup}
We used a small and fast pretrained language model,
DistilBERT \cite{sanh2019distilbert} as the backbone classifier 
% for all the AL models on all the text classification tasks 
in our experiments.
% DistilBERT has the same general architecture as BERT. 
% It is a light transformer model trained by distilling BERT base. The model has 40\% less parameters than BERT and the operating speed is 60\% faster then BERT while retaining 97\% of BERT's performance. 
We fine-tuned DistilBERT on each dataset after each AL iteration with a random re-initialization \cite{frankle2018lottery},
 proven to improve the model performance over the use of incremental fine-tuning with the newly acquired samples \cite{gal2017deep}.
% we use DistilBERT to encode the text with 
The maximum sequence length was set to 128, and a maximum of 30 epochs was used in fine-tuning DistilBERT with early stopping \cite{dodge2020fine}.
% and enabled fine-tuning.
We used AdamW \cite{loshchilov2017decoupled} as the optimizer with learning rate 2e-5 and betas 0.9/0.999. 
% All models are trained in PyTorch \cite{paszke2017automatic}.
Meanwhile, the initial training and validation split contain only 20 and 6 samples respectively.
% mimicking the scenarios where we only have a labelled pool of very limited size to start with.

Each AL method was run for five times with different random number seeds on each
dataset. 
The batch size $B$ 
% (i.e., the number of samples acquired at each AL iteration) 
was set to \{1, 5, 10, 50, 100\}.
% we run each AL method for five times with five different random seeds. 
% We set up 20 and 6 random examples as the initial labelled training set and validation set to fine-tune on the DistilBERT model. 
% In each fine-tuning run, the model is trained for 30 epochs with early stopping using a validation set \cite{dodge2020fine}.  
% To reduce the dependency of the model performance on a specific initialization \cite{frankle2018lottery}, we reinitialize the model with all labelled samples after each acquisition. 
% At each active learning acquisition iteration, the fixed number of samples is selected from the unlabelled pool according to the different batch size \{1, 5, 10, 50, 100\} and merged with the previous samples in the labelled pool.
To compute the predictive distributions $\prob{\cdot|L,x}$ and $\prob{\cdot|L,(x,y),x'}$,
we borrowed the idea of deep ensembles \cite{Lakshminarayanan2017},
where we trained five DistilBERTs with randomly generated
train-validation splits (i.e., train/validation=$70/30\%$) 
based on an incrementally augmented labelled pool.
Thus each member of the ensemble has a different train/validation set, helping to diversify the ensemble.
We call this split process \emph{dynamic validation set} (aka \textit{Dynamic VS}).
With it,
% To save labelling cost, as always required in domains like medicine, 
% 
our implementation of CoreMSE and CoreLog does not rely on a separate large validation set.  %  as in \cite{}.
% Instead, we randomly split the augmented training set with the newly acquired samples into 
% training and validation sets for each ensemble model after each AL iteration.
% We call it Dynamic VS.
% \lan{
% The size of the estimation pool $X$ and the top fraction $T$ for Algorithm~\ref{alg-qrd}
% were varied across datasets according to the size of each individual dataset.
More detailed experimental settings are given in Appendix~B.

\begin{table}[!t]
  \caption{The average running time (in seconds) per single acquisition iteration with unlabelled pools of varied sizes,
  which were generated from AG NEWS.
    % the size of the unlabelled pool. 
    % \wray{is this "per batch" or "per full acquisition run , 26-300 data" or what?} 
    }
  \label{tab:table2}
  \centering
%   \small
  \begin{adjustbox}{width=\textwidth}
  \begin{tabular}{llllllllll}
    \toprule
    \# Unlabelled  & CoreMSE & CoreLog & Max-Ent & BALD & MOCU  & WMOCU &BADGE  &ALPS &Rand  \\
    \cmidrule(r){2-10}
    10k     & 65 &65 &61 &61 &65 &65 &76 & 62 & <1     \\
    25k     & 163 &163 &159 &159 &163 &163 &323 &283 & <1    \\
    50k     & 326 &326 &322 &322 &326 &327 &884 &573 & <1      \\
    100k    & 654 &654 &649 & 650 &658 &659 &2904 &1299 & <1   \\
    \bottomrule
  \end{tabular}
  \end{adjustbox}
  \vspace{-3mm}
\end{table}

\begin{figure}[!t]
\centering
    \includegraphics[width=0.9\textwidth]{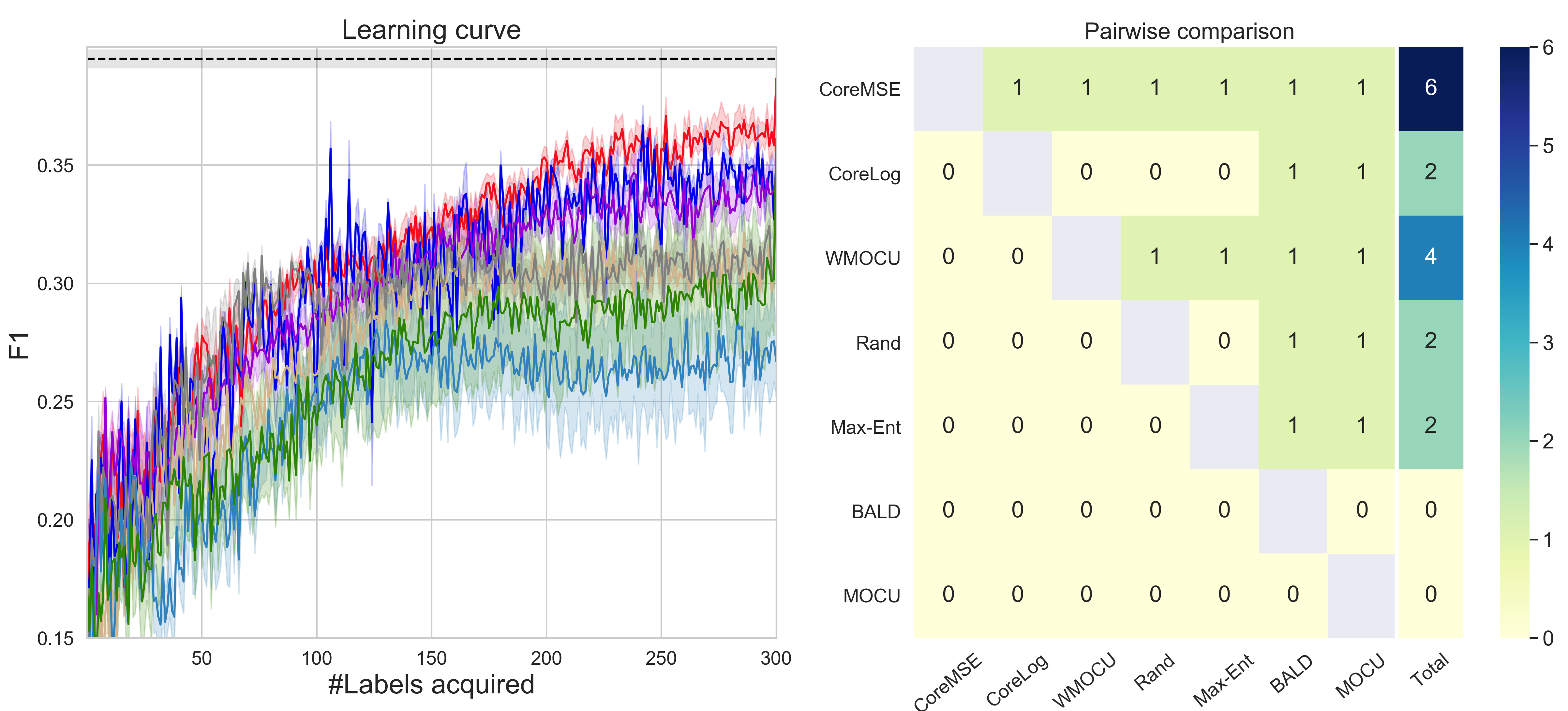}
    \includegraphics[width=1\textwidth]{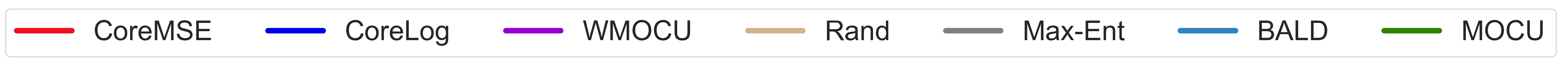}
     \caption{Performance on SST-5 dataset. The left half illustrates the learning curve, while the right half illustrates the matrix of paired comparisons. The dashline represents the performance of the backbone classifier trained on the entire dataset.
    %  CoreMSE outperforms all other uncertainty based methods.
    }
    \label{fig:uncertaintybasedAL_performance}
    % \vspace{-3mm}
% \end{figure}
% \begin{figure}[t]
%     \centering
%     \begin{subfigure}[b]{0.5\textwidth}
%         \centering
%         \includegraphics[width=\textwidth]{Figures/Pairwise Comparisons/b1_F1_pairwise.png}
%         \caption{F1-based pairwise comparison}
%     \end{subfigure}%
%     \begin{subfigure}[b]{0.5\textwidth}
%         \centering
%         \includegraphics[width=\textwidth]{Figures/Pairwise Comparisons/b1_Acc_pairwise.png}
%         \caption{Accuracy-based pairwise comparison}
%     \end{subfigure}
%     \caption{Pairwise comparisons on PUBMED and SST-5. 
%     % Each number in the matrices represents the number of times the corresponding method in the row beats the method in the column. 
%     The maximum value is two based on the number of datasets tested. 
%     % The number of the last column indicates the total winning times than the other methods. 
%     The larger value is the better.
%     }
%     \label{fig:uncertainty_pc_all}
    \vspace{-5mm}
\end{figure}

All experiments were run on 8 Tesla 16GB V100 GPUs.
For each AL method, we computed the average running time per one acquisition iteration 
as the total running time divided by the total number of iterations (i.e., 10).
% and total experimental runtime was about 228 hours. \wray{total runtime was 2,034 hours (fix me please).}
%
% Dataset= AG NEWS
% the estimation pool size =
% top fraction T value =
% batch size = 10
\autoref{tab:table2} summarizes as an example 
the running times of all the AL methods on AG NEWS with unlabelled pools of different size.
% \lan{
The batch size  $B$ was set to 50, top fraction $T$ to 50\% and the size of X to 600.
% }
% shows the average runtimes per one acquisition iteration for all the eight AL methods. 
% Each average runtime was computed as  
% For all AL acquisition strategies except random, the runtimes are determined according to the acquisition function running on the output of the DistilBERT model. 
Except for BADGE and the random baseline, the runtimes of the other methods increase nearly linearly
with the size of the unlabelled pool.
% Max-Entropy and BALD require the least amount of time to acquire \lan{XXX} samples,  
% The time used by CoreMSE, MOCU, and WMOCU is almost identical. 
% However, \lan{the running time of BADGE increase exponentially with the pool size}, which
% could attribute to the computation of gradient embeddings.
% a longer inference time than other AL acquisition strategies with the same batch size. Additionally, it consumes more time as the size of the acquired batch increases
% (batch size vs \# labelled size).

\textbf{Comparative performance metrics}
% To robustly analyse comparative performance,
We followed \citet{ash2019deep} to compute a pairwise comparison matrix
but use a counting-based algorithm \cite{shah2017simple},
as shown in the left of  \autoref{fig:uncertaintybasedAL_performance}.
% compare the significant difference in model performance across these datasets, we use the pairwise comparisons according to the method of \citet{ash2019deep}. Instead of penalising the score in the matrices, we use the counting-based algorithm \cite{shah2017simple}. 
The rows and columns of the matrix %in \autoref{fig:uncertaintybasedAL_performance} 
correspond to the AL methods used in our experiments. 
% described in Subsection~\ref{ssct-nobatch}. 
Each entry represents the outcome of the comparison between method $i$ and method $j$ over all datasets (D). 
Let $C_{i,j,d} = 1$ when method $i$ beats method $j$ on dataset $d$, 
and 0 otherwise.
Each cell value of the matrix is them computed as
$C_{i,j}=\sum_d^DC_{i,j, d}$.
% The outcome of pair comparison $C_{i,j}$ is given by $C_{i,j,d}$, which is 1 when strategy $i$ beats strategy $j$ on dataset $d$, 
% and 0 otherwise.
%\begin{equation}
%  C_{i,j} =
%    \begin{cases}
%      1 & \text{if strategy $i$ beats strategy $j$}\\
%      0 & \text{otherwise}
%    \end{cases}       
%\end{equation}
% The total quantity $C_{i,j}=\sum_d^D\mathbb{I}_{C_{i,j, d}=1}$.
% corresponds to the total number of pairwise comparisons won by strategy $i$ shown in last column of \autoref{fig:uncertaintybasedAL_performance}. 
To determine the value of $C_{i,j,d}$, 
we used a two-side paired $t$-test to compare their performance for 5  weighted 
F1 scores (or accuracy) at maximally spaced labelled sample sizes $\{l_{i,j,d}^1, l_{i,j,d}^2, ...,l_{i,j,d}^5\}$ 
from the learning curve. 
We compute the $t$-score as $t = \sqrt{5}\hat{\mu}/\hat{\sigma}$, where
$\hat{\mu}$ and $\hat{\sigma}$ are the usual sample mean and std.dev.
In Figure 1, we selected five samples at the different iterations according to the step size 50 in the experiment. For example, the first sample $l_{i,j,d}^1$ is chosen at the $50^{th}$ iteration, and the second sample $l_{i,j,d}^1$ is chosen at the $100^{th}$ iteration, etc. $\hat{\mu}=\frac{1}{5}\sum_{k=1}^5\left(l_{i,j,d}^k\right), \hat{\sigma} = \sqrt{\frac{1}{4}\sum_{k=1}^5\left(l_{i,j,d}^k - \hat{\mu}^2\right)}$. 
The $C_{i,j,d}$ is assigned to 1 if method $i$ beats method $j$ with $t$-score > 2.776 ($p$-value < 0.05). 
We accumulate the outcome of each pair comparison into the total quantity of each strategy (i.e., the ``Total'' column in the
matrix). 
The highest total quantity gives a ranking over the AL methods. 
We also report the learning curves of all the AL methods with both weighted F1 score and accuracy.

\subsection{Model performance: learning curves and comparative comparisons}\label{ssct-nobatch}

% In the no-batch scenario of uncertainty-based AL,
\textbf{Active learning with batch size one} We first compared our CoreMSE and CoreLog based on Algorithm~\ref{alg-qr} to the baselines on
% for classic uncertainty active learning strategies using 
the PUBMED and the SST-5 datasets
to demonstrate how those methods perform particularly on a hard classification setting
where classes are imbalanced.
% small imbalance datasets in \autoref{tab:table1}. 
% \autoref{fig:uncertaintybasedAL_performance} shows the performance of our methods compared to the other active learning strategies on SST-5. 
The learning curve sitting in the left of \autoref{fig:uncertaintybasedAL_performance}  shows
CoreMSE, CoreLog and WMOCU outperform all the other methods considered,  we attribute to 
their estimation of uncertainty being better related to  classification accuracy.
Among these three methods, our CoreMSE performs the best in term of F1 score.
The matrix at the right of
\autoref{fig:uncertaintybasedAL_performance} then presents a statistical summary of comparative performance.
% that CoreMSE's F1 performance is higher than other strategies. 
CoreMSE has the highest total quantity which further confirms its effectiveness in acquiring informative samples in AL. 
% significantly outperforms all other uncertainty based methods on the SST-5 dataset.
% \autoref{fig:uncertainty_pc_all} shows the pairwise comparisons computed based on either F1 scores
% or accuracy on both PUBMED and SST-5. CoreMSE generally outperforms all the baselines.
More results on SST-5 and PUBMED, 
including both learning curves and comparative matrices are reported in Appendix~C. 

% \begin{figure}[!t]
% \centering
%     \includegraphics[width=1\textwidth]{Figures/Learning Curves/b50_F1.png}
%     \includegraphics[width=1\textwidth]{Figures/Learning Curves/legend_diversity.png}
%     \caption{Learning curves of batch size 50 for PUBMED, IMDB, SST-5 AND AG NEWS.}
%     \label{fig:b50_f1_lc}
% % \end{figure}
% % \begin{figure}[!t]
% %     \centering
%     \begin{subfigure}[t]{0.5\textwidth}
%         \centering
%         \includegraphics[height=2in]{Figures/Pairwise Comparisons/Total_F1_pairwise.png}
%         \caption{F1-based pairwise comparison}
%     \end{subfigure}%
%     ~ 
%     \begin{subfigure}[t]{0.5\textwidth}
%         \centering
%         \includegraphics[height=2in]{Figures/Pairwise Comparisons/Total_Acc_pairwise.png}
%         \caption{Accuracy-based pairwise comparison}
%     \end{subfigure}
%     \caption{Pairwise comparison matrices of batch active learning strategies. 
%     % Each number in the matrices represents the number of times the corresponding method in the row beats the method in the column. The maximum value is four based on the number of datasets tested. The number of the last column indicates the total winning times than the other methods. The higher value is better.
%     }
%     \label{fig:divercity_sum_pc}
%     \vspace{-5mm}
% \end{figure}

\begin{figure}[!t]
\centering
    \includegraphics[width=1\textwidth]{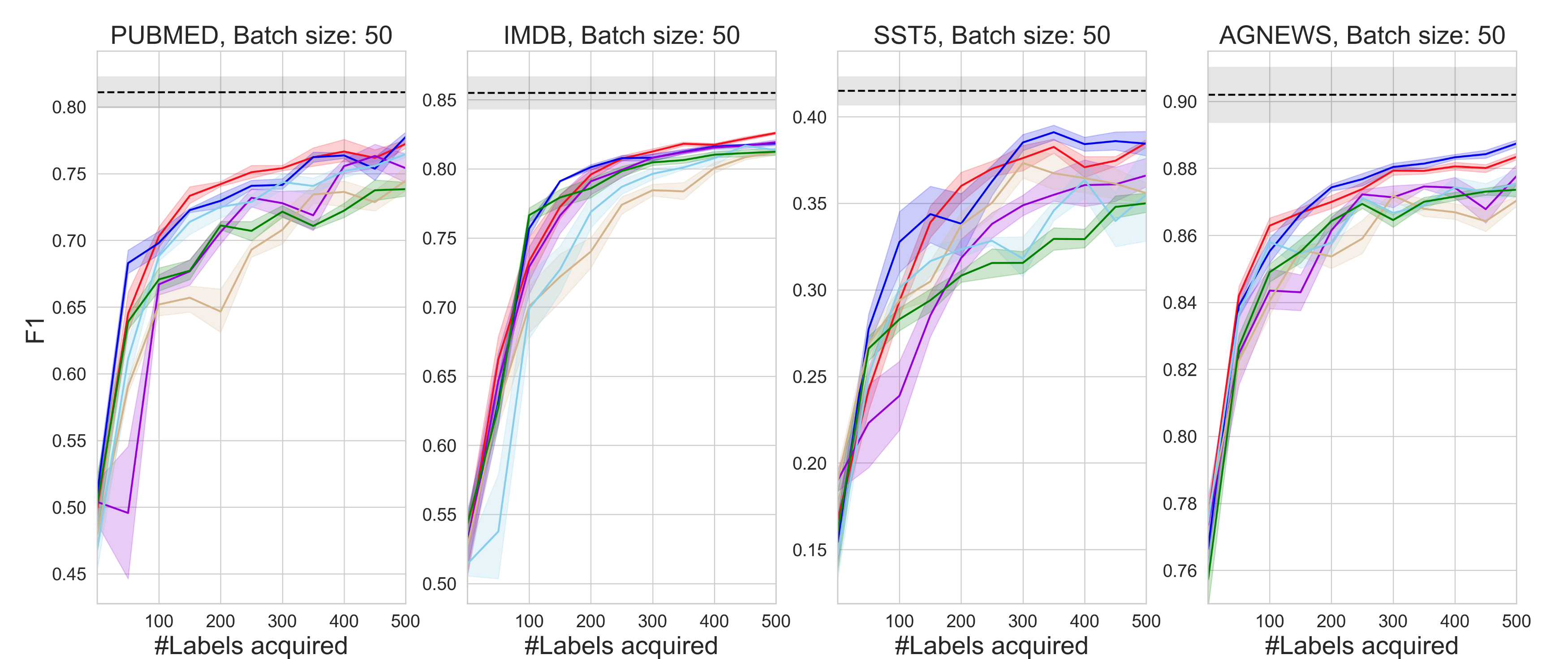}
    \includegraphics[width=1\textwidth]{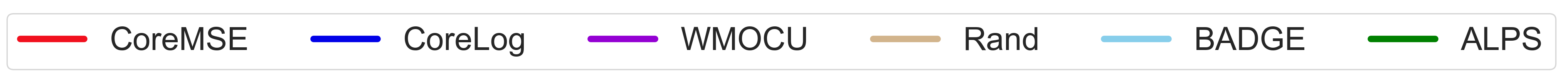}
    \caption{Learning curves of batch size 50 for PUBMED, IMDB, SST-5 and AG NEWS. The dashline represents the performance of the backbone classifier trained on the entire dataset.}
    \label{fig:b50_f1_lc}
% \end{figure}
% \begin{figure}[!t]
%     \centering
    \begin{subfigure}[t]{0.48\textwidth}
        \centering
        \includegraphics[width=\textwidth]{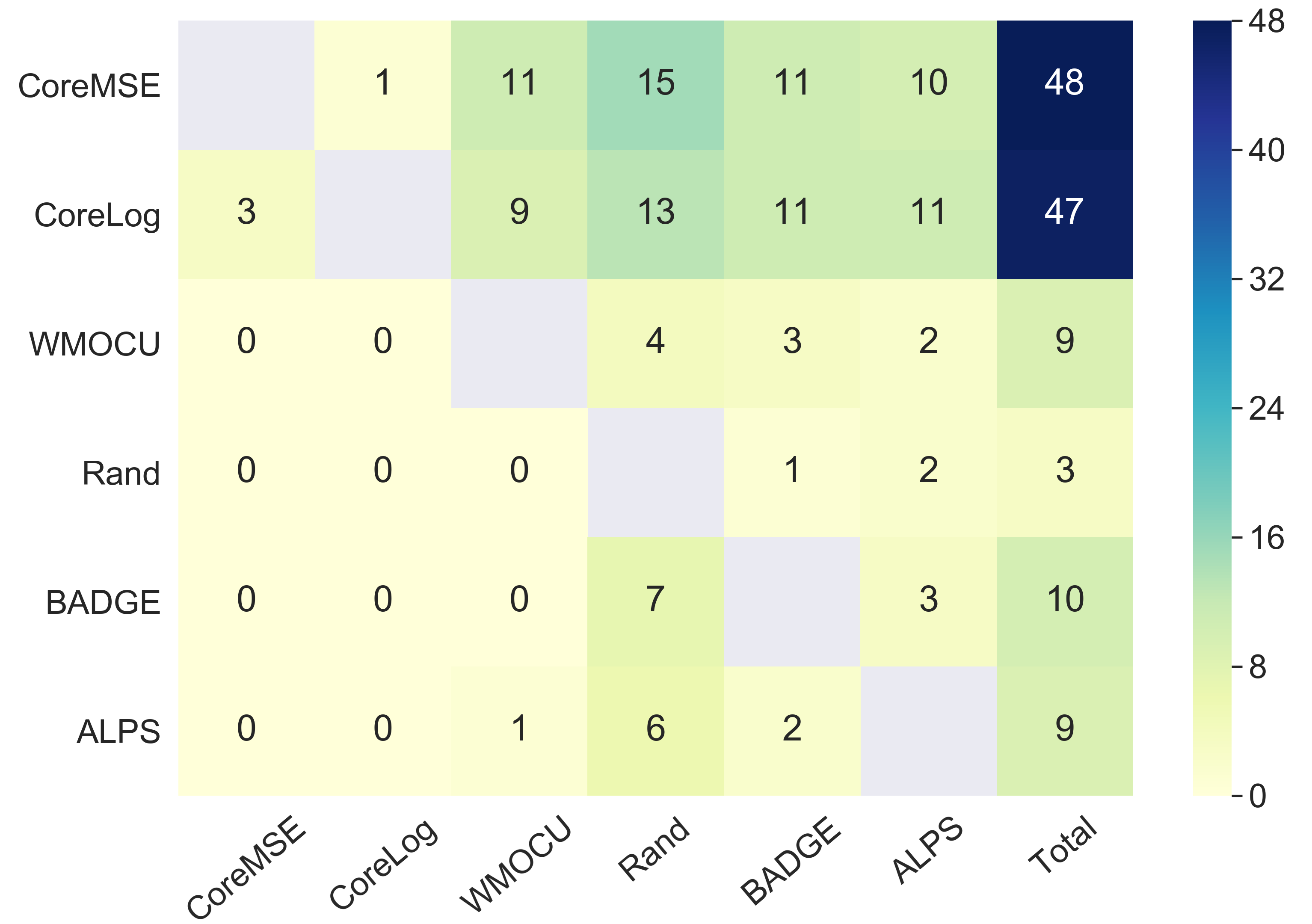}
        \caption{F1-based pairwise comparison}
    \end{subfigure}%
    ~ 
    \begin{subfigure}[t]{0.48\textwidth}
        \centering
        \includegraphics[width=\textwidth]{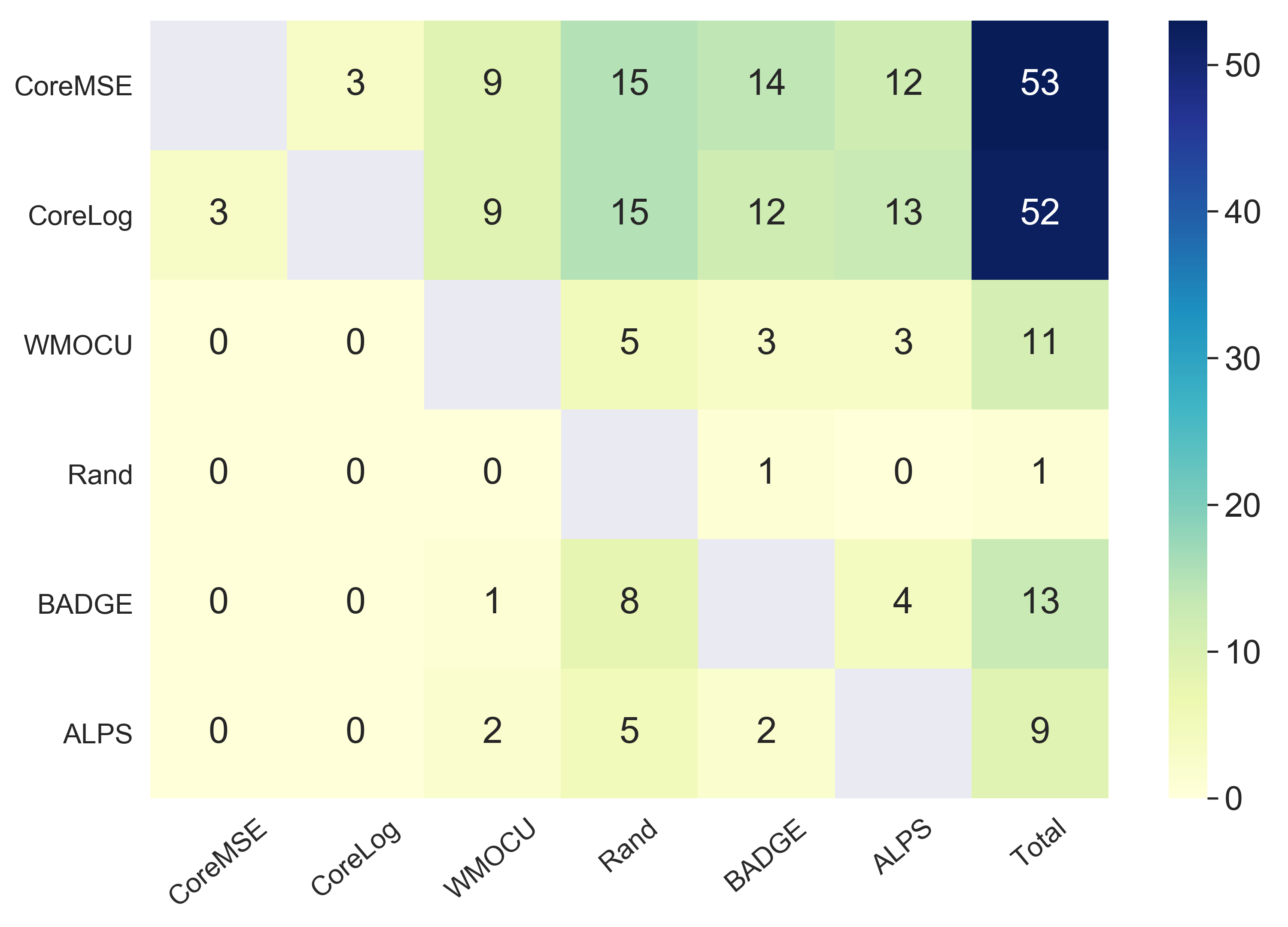}
        \caption{Accuracy-based pairwise comparison}
    \end{subfigure}
    \caption{Pairwise comparison matrices of batch active learning strategies. 
    % Each number in the matrices represents the number of times the corresponding method in the row beats the method in the column. The maximum value is four based on the number of datasets tested. The number of the last column indicates the total winning times than the other methods. The higher value is better.
    }
    \label{fig:divercity_sum_pc}
    \vspace{-5mm}
\end{figure}

\textbf{Batch active learning} 
We compared batch CoreMSE and batch CoreLog implemented based on Algorithm~\ref{alg-qrd} with
BADGE,
%%%%%%%%%%%%%%%%%%%%%%%%%%%%%%%%%%%%%%%%%%%%%%%%%%%%% uncomment "," above and 1131
 ALPS 
%%%%%%%%%%%%%%%%%%%%%%%%%%%%%%%%%%%%%%%%%%%%%%%%%%%%
and batch %Max-Entropy and 
WMOCU on the four datasets listed
in \autoref{tab:table1},
%%%WRAY:  this already mentioned under "Experiment setup"
% The batch size $B$ varied from \lan{5 to 100}.
% \footnote{We considered relatively small batch sizes, unlike many existing works \cite{},
% because in many real-world applications, like those in medicine, it is impractical and cost-inefficient to ask clinicians to annotate hundreds and thousands samples. Therefore, one of our focuses 
% is to improve AL performance with a limited annotation budget}.
We extended WMOCU with our Algorithm~\ref{alg-qrd} to build its batch counterpart.
Specifically, we generated $vec_x$ using its point-wise error estimates, i.e., 
% the weighted version of  Eq~\eqref{eq-q-mocu} (see 
Eq~(10) in \cite{ZhaoICLR21}. 
% For Max-Entropy we selected the top-$B$ ranked unlabelled samples.
The random baseline selected $B$ unlabelled samples randomly.
Here we present the results derived with $B=50$ as an example.
More comprehensive results with different batch sizes, including accuracy,
can be found in Appendix~C.
% We evaluate the performance of the batch versions of the  methods over the imbalanced and balanced datasets in . 
% Our methods CoreMSE and CoreLog have higher F1 and Accuracy than the baselines of diversity based methods.
% To demonstrate the  methods' performance against baselines, 
% we convert CoreMSE, CoreLog, WMOCU and MOCU to a batch version by 
% applying the kmeans clustering algorithm based on the vector of pointwise estimated scores or classifier errors,
% as described in Subsection~\ref{ssct:algos}. 
% Additionally, we compare them with the recently published method BADGE. 
% All the experiments are conducted with different batch sizes on the four datasets.
 
The learning curves in \autoref{fig:b50_f1_lc} show that batch CoreMSE and CoreLog almost always outperform the other AL methods as
the number of acquired samples increases. 
Batch WMOCU devised with our batch algorithm compare favourably 
% with BADGE that uses gradient embeddings to increase batch diversity.
%%%%%%%%%%%%%%%%%%%%%%%%%%%%%%%%%%%%%%%%%%%% replace 1163
with BADGE and ALPS that use gradient/surprisal embeddings to increase batch diversity.
%%%%%%%%%%%%%%%%%%%%%%%%%%%%%%%%%%%%%%%%%%%%
These results suggest that selecting the representative samples from clusters learned with vectors of expected change in scores 
can better improve batch diversity, leading to an improved AL performance. 
% \lan{Section~\ref{}} presents detailed ablation study on batch diversity. 
% Notably, clustering the estimated error embedding performs significantly better than gradient embedding clustering. 
Comparing CoreMSE/CoreLog with WMOCU further shows
the advantage of BEMPS.
% having a direct impact  on the classification error.
Moreover, the performance differences between our methods and others 
on PUBMED and SST-5 indicate that
% In other words, 
batch CoreMSE and CoreLog can still achieve good results 
when the annotation budget is limited in those imbalanced datasets.
% on imbalanced dataset, 
% which indicates that 
% they are capable of dealing with the issue of limited annotation budget faced by the imbalanced datasets.
% the largest improvement were observed on the imbalanced dataset. 
% For SST-5 and PUBMED datsets, the difference in F1 scores between our methods and WMOCU is even greater than the difference in accuracy. These findings show that CoreMSE and CoreLog can indeed improve results when the annotation budget is limited for the imbalanced datasets.  
% %  with a low prior for positive instances.
% Similarly, we use the pairwise comparisons to summarise all experiments on the varying batch size across all datasets. 
We also created four pairwise comparison matrices for different batch sizes using either F1 score or accuracy.
\autoref{fig:divercity_sum_pc} show the sum of the four matrices, summarizing 
the comparative performance on the four datasets.
The maximum cell value is now $4\times 4 = 16$. 
In other words,
if a method beats another on all the four datasets across the four different batch sizes,
the corresponding cell value will be 16.
% sum them together to finalise the ranking for all the methods. 
Both matrices computed with F1 score and accuracy respectively 
show both CoreMSE and CoreLog are ranked higher than the other methods.
% winning 70\% of 80 cases ($4\times 4 \times 5$) 
% with statistically significant improvement. 
% WMOCU and BADGE performs better than the other methods. The matrices with varying batch sizes in Appendix~\ref{app:exps} exhibit a similar trend.
The observations discussed above are also consistent across different batch sizes.
% Learning curves of all vary batch sizes show that CoreMSE and CoreLog outperform other methods over all datasets (see Appendix~\ref{app:exps}).

\begin{figure}[!t]
\centering
    \includegraphics[width=0.9\textwidth]{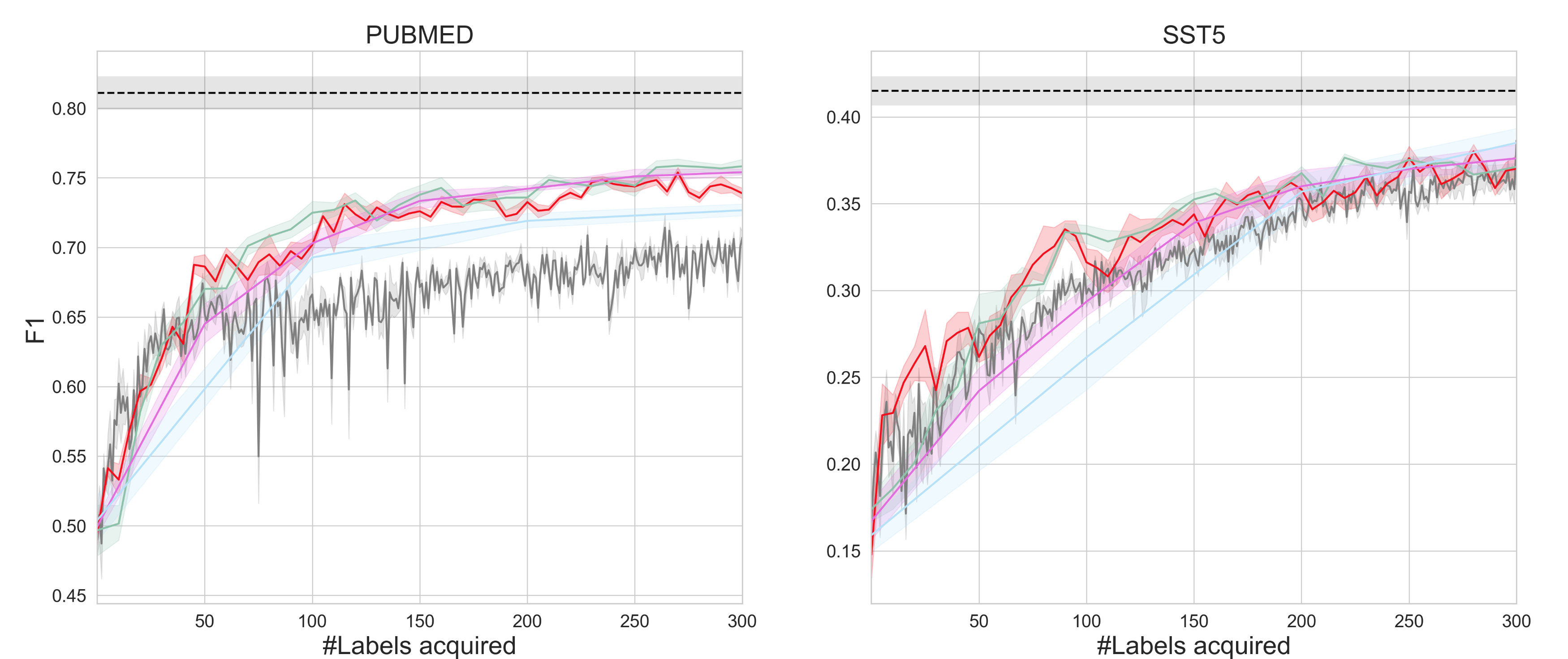}
    \includegraphics[width=1\textwidth]{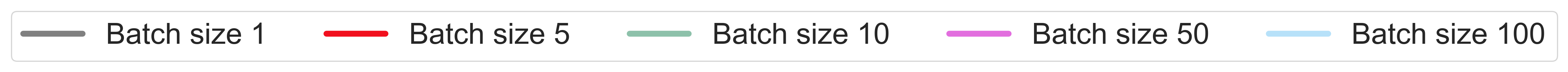}
    \caption{Learning curves of batch size 1, 5, 10, 50 and 100 for CoreMSE}
    \label{fig:lc_all_batchsize_coremse}
        \includegraphics[width=1\textwidth]{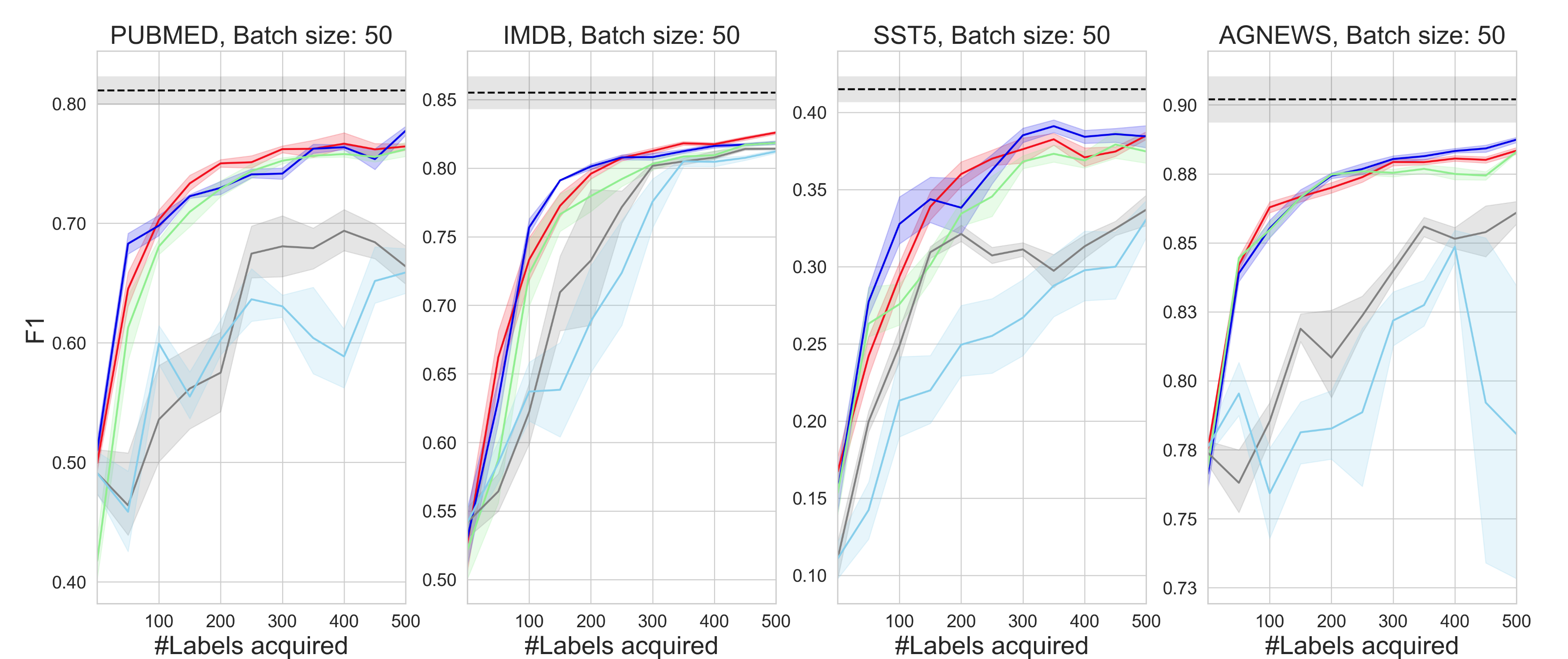}
    \includegraphics[width=1\textwidth]{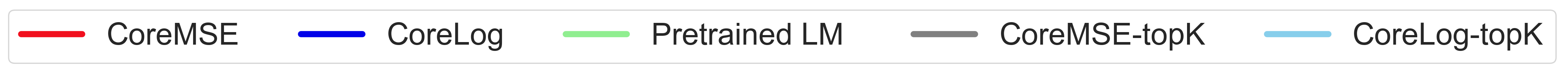}
    \caption{Learning curves of the model training with diversity. The dashline represents the performance of the backbone classifier trained on the entire dataset.}
    % \caption{Learning curves of different approaches used to generate a batch.}
    \label{fig:b50_diversity}
    % \includegraphics[width=\textwidth]{Figures/Ablation/Validation/b50_mixture_vs_F1.png}
    % \includegraphics[width=1\textwidth]{Figures/Ablation/Validation/legend_b50_mixture_vs.png}
    % \caption{Learning curves of the model training with a dynamic VS, constant VS,  a fixed \# epochs without VS, 1000 labels fixed length VS for CoreMSE}
    % \label{fig:b50_mixture_vs_coremse}
    \vspace{-6mm}
\end{figure}

\subsection{Ablation Studies}
\label{sec:ablation}

\textbf{Batch size}
% \textcolor{red}{The section reports our experiments evaluating the batch method used.}
% We demonstrate the occurrences of active learning algorithms in terms of batch size and dataset in this section using examples of learning curves.
In \autoref{fig:lc_all_batchsize_coremse}, we plotted the learning curves of batch CoreMSE with different batch sizes (i.e., 
$B \in\{1, 5, 10, 50, 100\}$) on PUBMED and SST-5 as an example.
The curves demonstrate that 
% batch CoreMSE (i.e., $B > 1$) performs better than CoreMSE (i.e., $B=1$).
% when the uncertainty approach is enhanced with diversity. 
the performance of smaller batch sizes (5 or 10) is 
superior to that of large batch sizes (50 or 100), especially in the early rounds of training,
but surprisingly also superior to that of the no-batch case.
% Given a batch size $B$,
The no-batch case need to perform multiple one-step-look-ahead optimising acquisitions (Algorithm 1 and Algorithm 2) sequentially in order to acquire $B$ samples whereas the batch case only perform the acquisition at once, heuristically, incorporating a form of diversity based directly on the error surface (Algorithm 1 and Algorithm 3).
This phenomena is also seen with BatchBALD \cite[see Figure~4]{KAJvAYG2019}.
Thus we can conclude that the one-step-look-ahead of
Equation~\eqref{eq-dQ} is only greedy, and not optimal.
%While the importance of diversity was illustrated in \citet{zhou2020understanding},
% they used larger batch sizes in their experiments.
% a phenomena also seen with BatchBALD \cite{KAJvAYG2019}.  
% While this was also the subject of a detailed
% computational analysis \cite{zhou2020understanding}, 
% they did use larger batch sizes in their experiments.
More results of 
% In addition, we plot the learning curves 
Batch CoreMSE and CoreLog on PUBMED, SST-5, AG NEWS and IMDB can be found in Appendix~B.
% in \autoref{fig:lc_b51050100_coremse} and \autoref{fig:lc_b51050100_corelog} (See Appendix~\ref{app:exps}), 
% +we can conclude the performance of smaller batch size consistently perform better then the large batch size. 
In general, all the results shown in the learning curves prove that batch algorithms improve AL performance over the no-batch ones, and smaller batch sizes are more preferable to the large ones.

\textbf{Batch diversity}
% We perform an ablation study on the four datasets. 
To further study the effectiveness of Algorithm~\ref{alg-qrd},
we considered the following variants: 
1) Pretrained\_LM: Instead of using the expected changes of scores to represent each unlabelled sample $x$, 
we used the embedding generated by the last layer of DistilBERT in $k$-Means clustering, 
which is similar to the BERT-KM in \cite{yuan-etal-2020-cold};
and 2) CoreMSE\_top and CoreLog\_top: We simply chose the top-$B$ samples ranked by $Q_x$.
% we remove AL functions and use k-MEANS clustering to select the number of samples whose representations directly generated from the last layer of DistilBERT. 
% 2) No k-MEANS clustering: we only use the top points with the highest BALD and Maxmium Entropy acquisition scores based on batch size. 
% 3) Random: samples are uniformly sampled from the unlabelled pool based on the batch size.
\autoref{fig:b50_diversity} shows the results for all these variants. 
% Each module contributes to the model's final performance. In the first variant, we use k-MEANS clustering to select samples closest to the cluster centroid based on the representation generated by pretrain LM. 
% \lan{ % Need to rewrite based on the new plots.
% \autoref{fig:b50_diversity} shows that 
Our batch CoreMSE and CoreLog perform much better than the corresponding
CoreMSE\_top and CoreLog\_top, which showcases Algorithm~\ref{alg-qrd} can promote batch diversity
that benefits AL for text classification.
The performance difference between Pretrained\_LM and batch CoreMSE/CoreLog
indicates that representing each unlabelled sample as vector of expected changes in scores (i.e., Equation~\eqref{eq-DQS}) is
effective in capturing the information to be used for diversification.

% \lan{
% Moreover,  
% clustering with embeddings generated by DistilBERT can also give us quite competitive 
% performance, which is inline with the findings in  \cite{yuan-etal-2020-cold}.
% }
% 
%Lan: please correct me if I am wrong.
%
%
% we can consider adding the following argument after Dad finish the experiments with B = 10:
% Pretrained LM highly relies on the quality of the fine-tuned DistilBERT, which can be easily affected
% by the size of training set and if the classes are imbalance.
% 

% the pretrained LM with CoreMSE and CoreLog outperforms the pretrained LM without using active learning algorithms. Our second variant emphasizes the value of diversity in AL algorithms. CoreMSE and CoreLog augment the k-MEANS clustering method to minimise the distance between the centroid and the actual representation in the unlablled pool by counting the expected error estimation. The uncertainty methods BALD and Maximum entropy are incapable of matching the performance of our model. Their performance on all datasets is significantly less than our model's. Finally, we compare our methods against the fundamental baseline Random.  As a result, CoreMSE and CoreLog can efficiently perform better than the baselines.

\begin{figure}[!t]
\centering
        \includegraphics[width=1\textwidth]{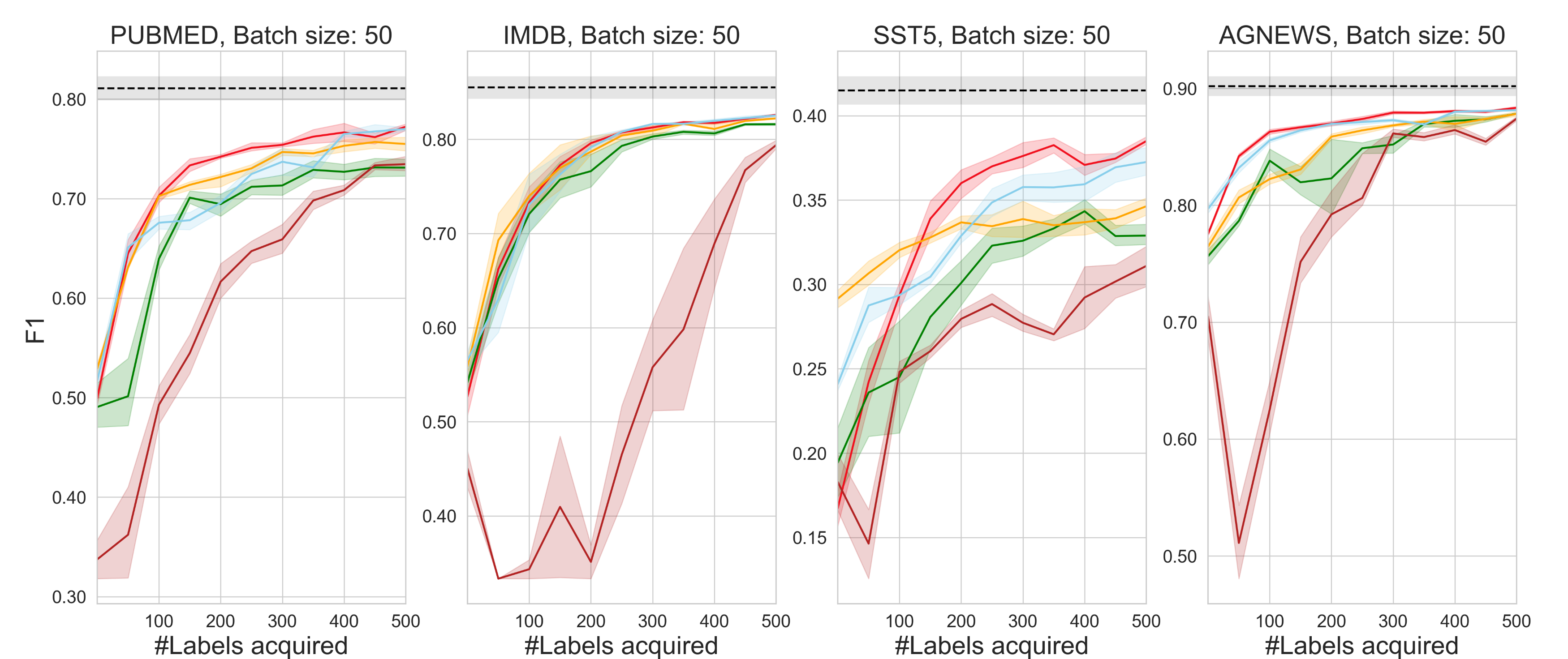}
    \includegraphics[width=1\textwidth]{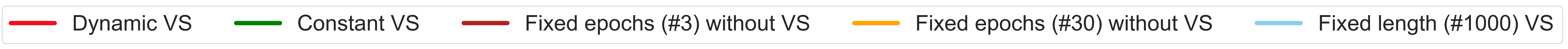}
    \caption{Learning curves of the model training with a dynamic validation set, constant validation set, fixed \# epochs without validation set, fixed length \# labels validation set for CoreMSE. The dashline represents the performance of the backbone classifier trained on the entire dataset.}
    \label{fig:b50_mixture_vs_coremse}
    \vspace{-6mm}
\end{figure}

\textbf{Dynamic VS}
% \textcolor{red}{The section reports our experiments evaluating the validation set method used.}
% To increase the training efficiency and model's performance, 
% We adopted a dynamic approach to generate validation sets for individual ensemble models.
% Difference 
% in order to increase both model efficiency and performance for AL methods. 
We studied how the dynamic VS impacts the ensemble model training by comparing 
batch CoreMSE with dynamic VS to
% The full batch CoreMSE was then compared with 
its following variations:
1) \textit{3/30 epochs without VS}: ensemble model training without VS and each model was trained for 3 or 30 epochs \cite{dodge2020fine},
2) \textit{Fixed length (\#1000) VS}: a pre-fixed validation set with 1000 labelled samples separate from the labelled pool, used in some existing AL empirical work;
3) \textit{Constant VS}: a variant of dynamic VS where one random split was generated after each AL iteration and then shared by all the ensemble models.
% the dynamic validation set approach's model performance to that of two other approaches: 
% variant one uses validation set after a fixed number of epochs, 
% variant two uses instead a fixed validation set, as most existing AL methods do.
% The results shown in \autoref{fig:b50_mixture_vs_coremse} highlight
% the need of validation set to learn descent classifiers for AL.
% To demonstrate CoreMSE's performance indifference between dynamic validation and fixed epochs without validation, we conduct experiments on the four datasets using five different scenarios for the batch size 50 CoreMSE method. All the scenarios are shown in \autoref{fig:b50_val_noVS_coremse}. We can see the performance of the dynamic validation set approach outperforms other scenarios which the model trains without validation set. It is thus evident that the use of validation set approach for the model training is obviously better than the model training without validation. This highlights the need to fine-tune the model with the validation set in AL that effectively performs for the each acquisition.
% \lan{% To be done after David generates the figure
\autoref{fig:b50_mixture_vs_coremse} shows the learning curves of those compared AL methods.
Dynamic VS gains an advantage after the third acquisition iteration
on PUBMED and SST-5, the first iteration on AG NEWS.
It is not surprising to see that 30 epochs without VS and Fixed length VS performs
better in the early acquisition iterations, since they used the whole augmented labelled pool in training DistilBERT, 
whereas CoreMSE used 70\%.
But choosing the number of epochs without a validation set is simply heuristic
otherwise.
Also Fixed length VS is midway between Constant VS and Dynamic VS, indicating the variability in ensembles inherent in the dynamic training sets is a source of improvement.
% (e.g., 3 epochs without VS v.s.~30 epochs without VS).
More ablation results are reported in Appendix~D.

\section{Conclusion}

We developed the BEMPS framework for acquisition functions for AL based
around strictly proper scoring rules \cite{doi:10.1198/016214506000001437}, or alternatively, Bregman divergences.
These are fundamental scores in statistical learning theory
and almost universally used when training neural networks,
unlike simple errors, thus make a solid basis.
In experiments we used mean squared error and log probability,
making new acquisition functions CoreMSE and CoreLog respectively.
For this, we developed convergence theory,
borrowing techniques from \cite{ZhaoICLR21},
that we also extended to the earlier BALD acquisition function. 
A primary limitation of the theory is that it does not support the use of unlabelled data within the learning algorithm, for instance, as would be done by state of the art semi-supervised learning methods.
Given that semi-supervised learning and AL really address a common problem,
this represents an area for future work for our techniques.
But note recent high performing batch AL uses semi-supervised learning, thus we excluded them from comparison.

For more efficient and effective evaluation of the new BEMPS based 
acquisition functions,
we developed two techniques to improve performance,
a batch AL strategy naturally complement to our BEMPS algorithms, and a dynamic validation-ensembling hybrid that 
generates high scoring ensembles but does not require a separate large labelled dataset.
%  "societal impact" comment
While this ensembling increases computational demand, it is only done for small datasets.
These were both tested in isolation and shown to perform well.
Though, interestingly, 
batch BEMPS works better than no-batch BEMPS, also seen for BatchBALD.  This suggests that the one-step-look-ahead can be improved, and additional theory is needed.
Finally, we followed some of the strong evaluation standards set in earlier
research, but we kept to small labelled set sizes in keeping with text annotation practice,
testing our approach against WMOCU, BADGE, BALD and baselines.
Using mean squared error and log probability as the scoring rules yielded
consistently high-performing AL on a variety of data sets.

\section*{Acknowledgements}
%
%  Turning Point, DARPA, MRFF (Lan)
The work has been supported by the Tides Foundation through Grant  1904-57761, as part of the Google AI Impact Challenge, with Turning Point.
Wray Buntine's work was also partly supported by DARPA’s Learning with Less Labelling (LwLL) program under agreement FA8750-19-2-0501.

%%
%%  not in pagesize
{\small
\bibliographystyle{plainnat}
\bibliography{neurips}
}

\end{document}